\documentclass[10pt,twocolumn,letterpaper]{article}
 
\usepackage{cvpr}              

\usepackage[accsupp]{axessibility}

\usepackage[ruled,linesnumbered]{algorithm2e}
\usepackage{graphicx}
\usepackage{animate}
\usepackage{multirow}
\usepackage{nth}
\usepackage[dvipsnames]{xcolor}

\definecolor{cvprblue}{rgb}{0.21,0.49,0.74}
\usepackage[pagebackref,breaklinks,colorlinks,citecolor=cvprblue]{hyperref}

\def\paperID{3659} 
\def\confName{CVPR}
\def\confYear{2024}

\title{Make-Your-Anchor: A Diffusion-based 2D Avatar Generation Framework}

\author{Ziyao Huang$^1$,~Fan Tang$^1$,~Yong Zhang$^2$,~Xiaodong Cun$^2$,~Juan Cao$^{1*}$,~Jintao Li$^1$,~Tong-Yee Lee$^3$\\
$^1$ Institute of Computing Technology, Chinese Academy of Sciences
 ~ $^2$ Tencent AI Lab  \\
 ~ $^3$ National Cheng-Kung University\\
 {\tt\small \{huangziyao19f,~tangfan\}@ict.ac.cn 
 \{zhangyong201303,~vinthony\}@gmail.com} \\
 {\tt\small \{caojuan, jtli\}@ict.ac.cn tonylee@mail.ncku.edu.tw}
}

\begin{document}
 
\twocolumn[{%
\renewcommand\twocolumn[1][]{#1}%
\maketitle
\begin{center}
    \captionsetup{type=figure}
    \includegraphics[width=0.96\linewidth]{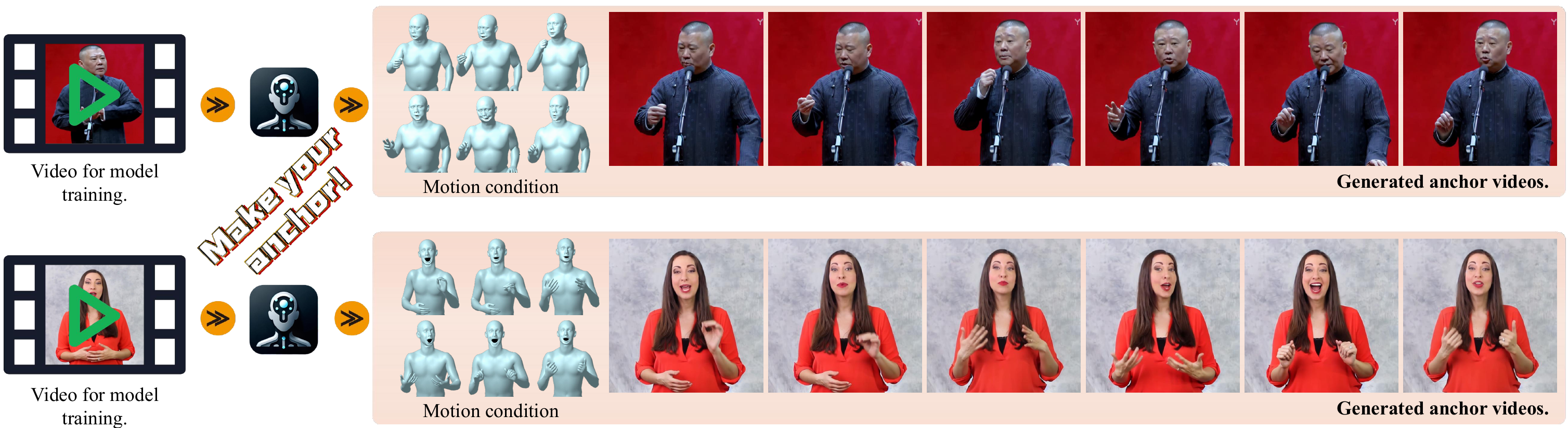}
    \caption{We propose Make-Your-Anchor, a diffusion-based 2D avatar generation framework, to map the animatable human 3D mesh sequence into realistic human video. By combining with motion capturing or audio-to-motion methods, our framework could achieve anchor video auto-generation with temporal and lifelike outcomes.}
   \label{fig:teaser}
\end{center}%
}]

\renewcommand{\thefootnote}{}
\footnotetext{\textsuperscript{*}Corresponding author: Juan Cao.}

\begin{abstract}
Despite the remarkable process of talking-head-based avatar-creating solutions, directly generating anchor-style videos with full-body motions remains challenging.
In this study, we propose Make-Your-Anchor, a novel system necessitating only a one-minute video clip of an individual for training, subsequently enabling the automatic generation of anchor-style videos with precise torso and hand movements. Specifically, we finetune a proposed structure-guided diffusion model on input video to render 3D mesh conditions into human appearances.
We adopt a two-stage training strategy for the diffusion model, effectively binding movements with specific appearances. 
To produce arbitrary long temporal video, we extend the 2D U-Net in the frame-wise diffusion model to a 3D style without additional training cost, and a simple yet effective batch-overlapped temporal denoising module is proposed to bypass the constraints on video length during inference.
Finally, a novel identity-specific face enhancement module is introduced to improve the visual quality of facial regions in the output videos.
Comparative experiments demonstrate the effectiveness and superiority of the system in terms of visual quality, temporal coherence, and identity preservation, outperforming SOTA diffusion/non-diffusion methods. 
Project page: \url{https://github.com/ICTMCG/Make-Your-Anchor}.
\end{abstract}

\section{Introduction}
\label{sec:intro}
Automatically generating 2D human videos with vivid expressions, torsos, and gestures under various conditions such as audio, text, or motion has wide applications in e-commerce, online education/conferencing, VR, etc. 
While AI-powered digital anchors offer certain benefits, such as around-the-clock availability and the ability to generate new content quickly, they also raise significant quality concerns. 

A common practice in popular 2D avatar systems is pronounced specifically as GAN-based talking face generation approaches~\cite{prajwal2020wav2lip, wang2021facevid2vid, zhang2023sadtalker, pang2023dpe}, where only the lip/facial region has been edited.
This type of methods relies on the capability to modify faces based on GANs~\cite{karras2019stylegan, karras2020stylegan2, wang2023cvm_face}.
Despite the remarkable visual quality, these systems limit the anchor's degrees of freedom.
Beyond regional editing, researchers try to generate more natural digital avatars with the help of motion transfer~\cite{motiontransfer_zhao2022tpsmotion, motiontransfer_siarohin2021mraa, motiontransfer_siarohin2019fomm, motiontransfer_yang2022delving} and co-speech generation~\cite{cogesture_qian2021speechdrivetemplates, cogesture_zhou2022audiodriven-videomotiongraphs}.
By learning the mapping from motion/speech to a person's appearance, these methods could generate full-body talking videos while moving the torso and hands.
However, the GAN-based solution limits the visual quality of the generated videos in case of fine details or fast movement.

Recently, diffusion models~\cite{ho2020ddpm} have been proven to achieve promising generation quality on human image generation~\cite{rombach2022ldm}. 
Compositing with pose or depth condition~\cite{zhang2023controlnet, mou2023t2iadapter}, current text-to-image diffusion models, such as Stable Diffusion~\cite{rombach2022ldm}, can synthesize human body with a specific gesture, pose, and expression, and can keep identity by personalized fine-tuning methods~\cite{ruiz2023dreambooth, gal2022TI, kumari2023customdiffusion, hu2021lora}.
However, due to the randomness of the diffusion process, image diffusion models cannot be directly applied to generate temporal-consistent human videos.

Existing video diffusion models~\cite{ho2022imagenvideo, singer2022makeavideo}, proven to generate temporal-consistent videos with simple motion and nature scenes.
DreamPose~\cite{karras2023dreampose} modifies Stable Diffusion to be controlled by input image and pose and animates a still human image conditioned on a given motion sequence.
DisCo~\cite{wang2023disco}, which also adopts an image as input, applies pose and background ControlNet~\cite{zhang2023controlnet} to composite different motions and scenes with a pretraining strategy and achieves promising human dance generation results on generalization.
However, these existing diffusion-based approaches still leave it challenging to synthesize facial details, exact gestures, and temporal frames, which is insufficient for making vivid digital anchors.

In this study, we propose ``Make-Your-Anchor'', a diffusion-based 2D avatar generation system to create customized digital avatars.
Despite the impressive visual quality of diffusion-based approaches, generating human images/videos has been challenging, in terms of identity preservation and motion consistency.
To this end, we learn to link human appearance with parametric human representation.
We replace the text clip inputs with image appearances, and a ControlNet-style modulation network is proposed to bind the human appearance to the pose control sequence frame-by-frame.
After pertaining on the existing news anchor dataset~\cite{yi2023talkshow}, we finetune the diffusion model on a specific person with a one-minute video.
To generate consistent temporal videos of any duration, we propose batch-overlapped temporal denoising by calculating the frame-wise latent noise in a cross-frame fashion during the inference phase.
Furthermore, special attention was paid to the facial region by applying a novel inpainting-based enhancement operation.

With the frame-wise training and batch-wise inference, the proposed system can be performed within a practical time cost on a single 40G A100 GPU.
Powered by existing technological solutions for generating corresponding 3D human mesh with lips, gestures, and body motions conditioned on music~\cite{zhou2023letdance}, audio~\cite{yi2023talkshow} or motion capture~\cite{cao2017openpose}, the system could make AI-powered digital anchors with vivid expressions, torsos, and gestures. 
Compared with popular commercial software, which is accomplished by talking face generation, our system achieves natural human video generation with more flexibility.
The main contributions are as follows:
\begin{itemize}
    \item We propose ``Make-Your-Anchor'', a 2D avatar customized system, to generate practical and applicable digital anchors with vivid lips, expressions, gestures, and body actions.
    \item We propose a frame-wise motion-to-appearance diffusing to bind movements into appearance with a two-stage training strategy, and batch-overlapped temporal denoising paradigm to generate temporal-consistent human video with long duration.   
    \item Qualitative and quantitative experiments on ten anchors validate the effectiveness of the proposed system compared with SOTA GAN-based motion transfer and diffusion-based human video generation approaches.
\end{itemize}
\section{System Overview}
\label{sec:method.overview}
\paragraph{Setting.}
Despite the impressive progress in talking face~\cite{zhang2023sadtalker} or fashion video generation~\cite{karras2023dreampose}, creating human video from a single or few images limits the ability to apply these methods in real-world applications.
Talking head videos are restricted to face regions, while DreamPose could only output fashion videos with a limited movement range.
In this study, we formulate 2D avatar generation as a learning paradigm from a video of one identity.
The setting is similar to the traditional talking head methods or commercial services such as ZenVideo.
However, these methods only learn from facial regions and use body motions of the input video as ``templates'' which are looped in the generated videos. 
The core of our system is to learn a personalized diffusion model that could generate a human video in the same scenario as the input video.
To this end, we tune the diffusion model to ``bind'' on the input video with appearance conditioned on motion conditions.
To the best of our knowledge, the proposed system is the first to generate 2D human avatar videos that could be given full-body motions with high visual fidelity in terms of identity preservation and temporal consistency.

\paragraph{Input.}
Based on the above setting, our system requires an anchor-style source video $\mathbb{S}$ of one person for training.
The given video $\mathbb{S}$ for training could contain lip, body, and gesture movements in anchor style, and the video length should be longer than one minute. 
Unlike DreamPose or ControlNet with OpenPose, we utilize human 3D mesh rendered from SMPL-X parameters~\cite{pavlakos2019smplx} as motion conditions.
The input pose condition's smoothness and accuracy influence the output video's temporal consistency, and the 3D human mesh has more structural information about the motion to generate smooth video, especially in hand gestures.
For inference, a sequence of pose $\mathbb{P}=\{p_1, p_2, ..., p_n\}$ is given to guide the 2D avatar learned by diffusion models.
Notably, our system can be effectively combined with existing human motion sequence generation methods such as audio-to-motion generation~\cite{yi2023talkshow} or video-guided motion transfer~\cite{cao2017openpose}.
In this case, the input could be a clip of audio or motion video reference.
The final output is human motion video frames $\mathbb{X}=\{x_1, x_2, ..., x_n\}$ with temporal appearance.

\section{Methodology}

\begin{figure*}
  \centering
   \includegraphics[width=1.0\linewidth]{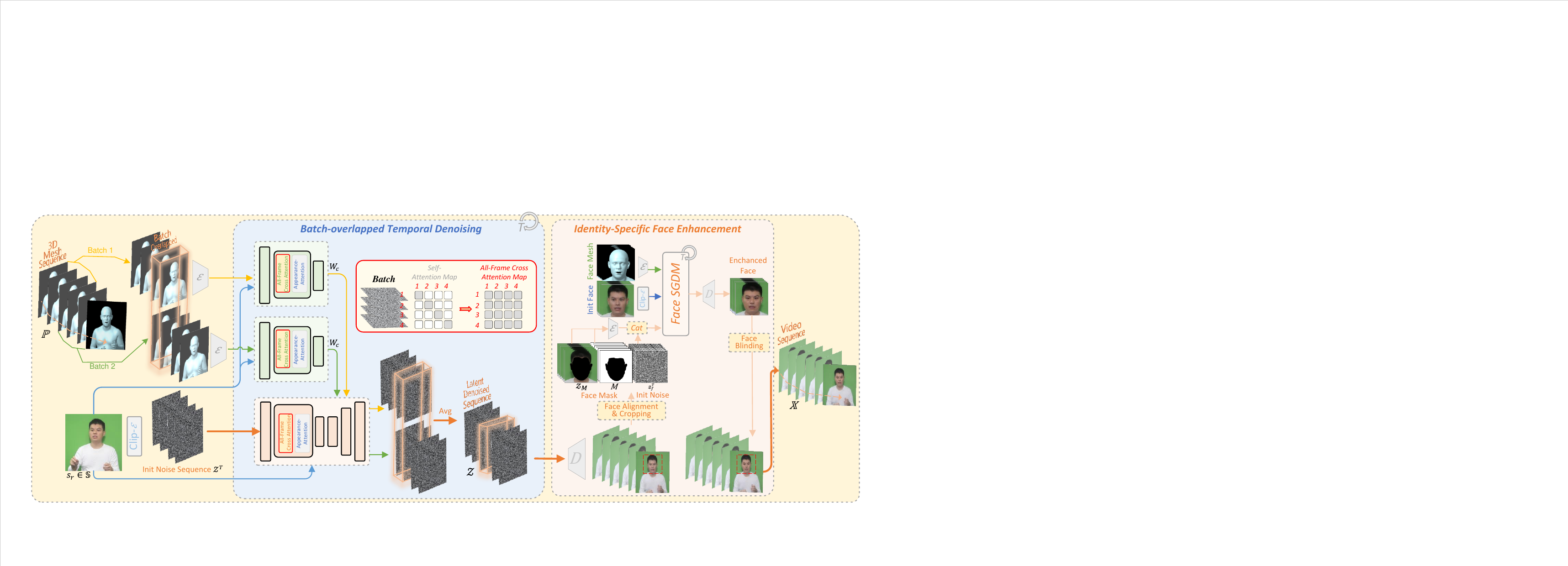}
   \caption{
   The inference pipeline of our system. An appearance condition and a 3D mesh sequence are inputted into the structure-guided diffusion, incorporating Batch-overlapped Temporal Denoising to accomplish video-level inference. Following the generation of arbitrary-length frame sequences, an inpainting-style module known as Identity-Specific Face Enhancement is utilized to enhance facial details.
   }
  \label{fig:pipeline}
\end{figure*}

Fig.~\ref{fig:pipeline} shows the network structure and the inference pipeline of the proposed system.
In this section, we start by introducing the diffusion-based network architecture to accomplish image-level appearance and body structure control in Sec.~\ref{sec:method.arch}.
Then we move on to the design of batch-overlapped temporal denoising in Sec.~\ref{sec:method.slindingwindow}, which is proposed to generate arbitrary long temporal consistent video without additional training efforts.
Finally, we propose an enhancement-by-inpainting module to improve the visual quality of facial regions in Sec.~\ref{sec:method.faceenhance}.

\subsection{Frame-wise Motion-to-Appearance Diffusing}
\label{sec:method.arch}
\paragraph{Preliminaries.}
Given an initial noise $\epsilon \sim \mathcal N(0, 1)$ and a condition $c$, the training loss of vanilla frame-wise diffusion models can be written as:
\begin{equation}
    \mathbb{E}_{s_i,c,\epsilon,t} [w_t || \hat{X}_{\theta} (\alpha_t s_i + \sigma_t \epsilon, t, c) -\sigma ||^2_2 ],
\end{equation}
where $\hat{X}_{\theta}$ is the diffusion model, $s_i$ is the $i_{th}$ frame in input source video $\mathbb{S}$, $\alpha_t$, $\sigma_t$, $w_t$ are hyperparameters, and $t$ is the diffusion timestep.
After training, the final output target frame $x_i = s_i^0$ is denoised from an initial noise $s_i^T$ with $T$ steps sampling.
In this study, we follow latent diffusion~\cite{rombach2022ldm} and use pretrained VAE encoder $\mathcal{E(\cdot)}$ to map the source frame $s_i$ into latent code ${z_i}$ and then conduct diffusing and denoising process in the latent space.
Finally, a pretrained VAE Decoder $\mathcal{D(\cdot)}$ maps the latent noise into pixel space. 

\begin{figure}
  \centering
   \includegraphics[width=1.0\linewidth]{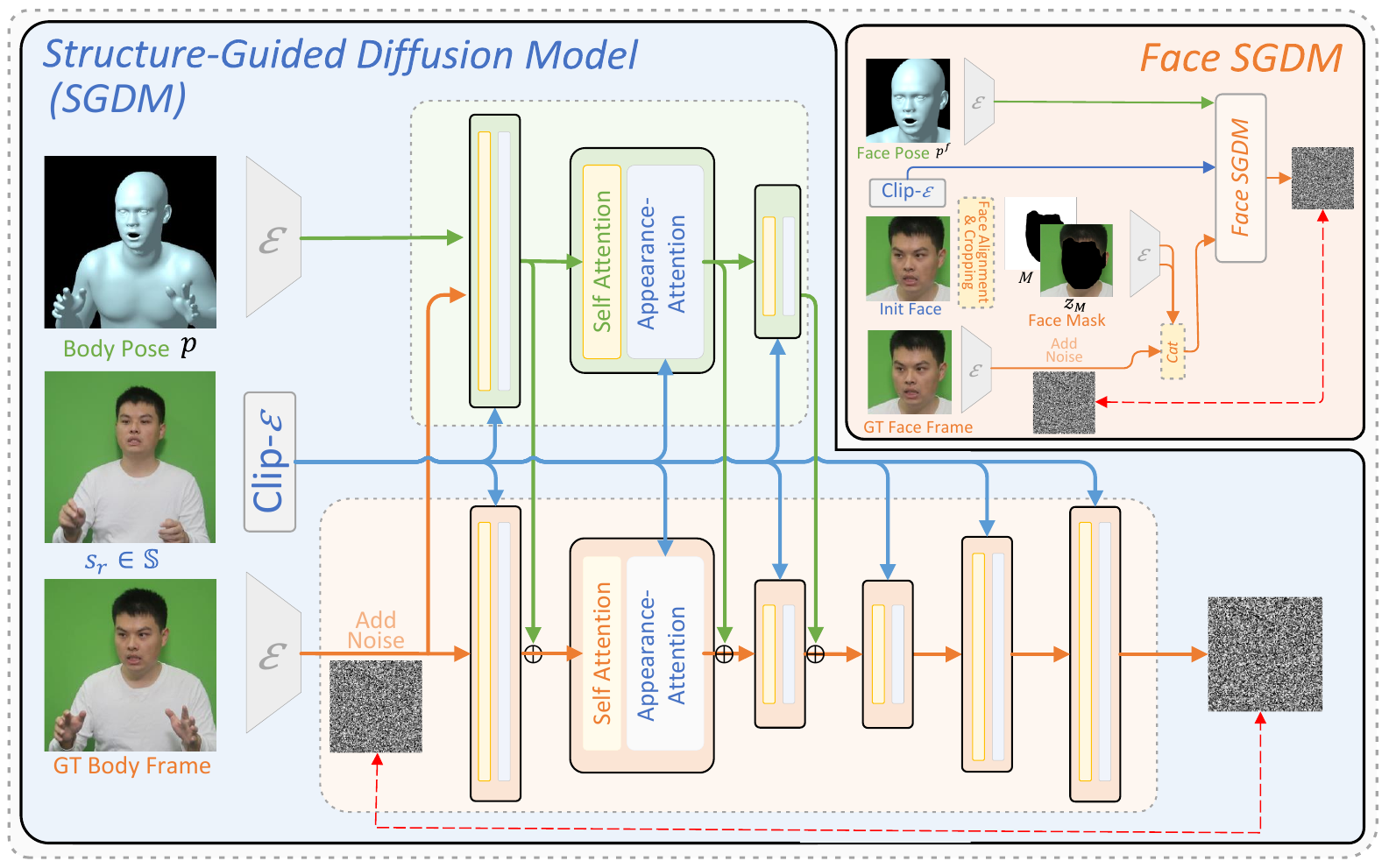}
   \caption{
   The network architecture of our proposed Structure-Guided Diffusion Model (SGDM) and Face SGDM.
   Our network achieves motion-to-appearance generation by embedding pose and appearance conditions into the pretrained diffusion model.
   }
  \label{fig:nn_struct}
\end{figure}

\paragraph{Structure-Guided Diffusion Model.}
We propose a structure-guided diffusion model (SGDM) to generate human videos under the control of 3D mesh conditions frame-by-frame, as shown in Fig.~\ref{fig:nn_struct}.
Different from ControlNet~\cite{zhang2023controlnet}, we embed the 3D mesh condition into the generation process to learn the corresponding mapping from pose $\mathbb{P}$ to target video frames $\mathbb{X}$.
An extra condition branch to the diffusion model's U-Net is copied from U-Net's encoder and a zero-convolution layer is added between the input condition image and encoder. 
The encoder's layer-wise feature maps are added on the U-Net's up layers:
\begin{equation}
    F_{up} = F_{up} + W_c \times ControlN(p_i),
\end{equation}
where $F_{up}$ is the feature map of U-Net's up layers, $ControlN(\cdot)$ is the additional module, $p_{i}$ is the $i_{th}$ 3D mesh in $\mathbb{P}$ and $W_c$ is a hyperparameter to weight condition.

However, adding a motion-control branch may still lead the diffusion model to lose control of the person's identity.
To solve this problem, we inject the appearance information of the same person by randomly choosing another source frame $s_r$ from $\mathbb{S}$ and then replace the text CLIP~\cite{radford2021clip} with image CLIP feature $c_{s_r}$ of $s_r$ to the cross-attention module of the latent diffusion's U-Net as well as the motion-control branch.
Specifically, we utilize the CLIP feature map before the global pooling layer, which provides more fine-grained appearance information.
After training, the generation of the human image can be abbreviated as:
\begin{align}
    z_i^0 &= \hat{X}_\theta( ControlN(p_i), c_{s_r}, z_i^T), \\
    x_i &= \mathcal{D}(z_i^0).
\end{align}

\paragraph{Two-Stage Training Strategy.}
\label{sec:method.twostagetrain}
We adopt the pretrained text-to-image Stable Diffusion V1.5~\cite{rombach2022ldm} as a basic model and build the proposed SGDM.
The training goal of our system is to bind the movement and appearance of the target identity by an ``binding'' style approach.
Therefore, we propose a two-stage training strategy, pre-training on multiple identities to enhance the models' ability for motion generation, and fine-tuning on a single identity to bind the movements to the appearance.
A publicly available anchor video dataset~\cite{yi2023talkshow} is used to train the new branch and tune the basic diffusion model using multiple identity data.
Then, we fine-tune the whole network based on the given one-shot videos of an arbitrary identity. 

To eliminate the interference of background on learning human information in the pre-training stage, we segment the foreground human body from the target images. 
During the fine-tuning stage on a single identity, we chose to fine-tune the model on the background and foreground together. 
This approach allows the model to quickly generalize to new individual identities while also remembering the background, thereby generating harmonious anchor videos.

\subsection{Batch-overlapped Temporal Denoising}
\label{sec:method.slindingwindow}
Adopting the two-stage train strategy to the proposed SGDM, our system could generate avatar videos frame-by-frame faithful to motion sequence.
However, due to the inherent randomness in the diffusion model's output, discontinuity still exists between frames.
Previous studies~\cite{khachatryan2023text2video, zhang2023controlvideo} have already demonstrated that extending self-attention in the image diffusion model to cross-frame attention enhances the consistency of generated videos. 
This study extends the 2D U-Net to a 3D U-Net to get a training-free video diffusion model $\hat{X}_\theta^V$. 
Specifically, we propose an all-frame cross-attention module to replace the self-attention in a data batch $\mathbb{B}$, which can be written as:
\begin{align}
    Attention_\mathbb{B}& := Softmax(\frac{Q_\mathbb{B} K_\mathbb{B}^T}{\sqrt{d}})\cdot V_\mathbb{B}, \\
    Q_\mathbb{B} &= W_Q \cdot Z^t,\\
    K_\mathbb{B} &=W_K \cdot Z^t,\\ 
    V_\mathbb{B} &=W_V \cdot Z^t,
\end{align}
where $Z^t=\{z_1^t, z_2^t, ... z_N^t\}$ is the input latent frames on diffusion timestep $t$ and $W_Q$, $W_K$, $W_V$ are learnable weights for attention modules.

Furthermore, we propose an overlapped temporal denoising algorithm to generate an arbitrary length of anchor video as illustrated in Algorithm~\ref{algorithm:temporal}.
Unlike existing methods that optimise the objective function between multiple batch noise~\cite{wang2023genlvideo} or use window attention~\cite{qiu2023freenoise}, we propose an effective and simple algorithm operated on multi-batch noise to achieve this goal.
We partition a long motion sequence input into multiple overlapped windows.
In each denoising step, the windows are fed one by one into our video diffusion model. 
Once all windows are handled, we normalize the overlapped noises between windows to ensure coherence, and then all video frames are denoised by the smoothing noises.
After all denoising steps are executed, the final results present a long temporal video.
We observe that this simple approach effectively extends the video diffusion model to generate videos of arbitrary length while keeping temporal consistency.

\subsection{Identity-Specific Face Enhancement}
\label{sec:method.faceenhance}
Generating a satisfied face from a holistic human generation is challenging.
From our perspective, this phenomenon arises due to the relatively diminutive size of the facial region within the entire image. 
With the training loss on holistic human image, the model encounters difficulty in fitting the small facial area.
We believe that a feasible solution is adapting the model to the key information of the face.
To address this issue, we opted for an inpainting-based approach with crop-and-blend operations to revising the facial region from the holistic generated body.
In detail, we modify the proposed SGDM to Face SGDM, as shown in the right part of Fig.~\ref{fig:nn_struct}.
The ready-to-revise face is cropped from the body generation, and the inpainting region is segmented via a facial mask $M$. 
We replace the input for SGDM's U-Net with a concatenation of facial mask $M$, masked face latent $z_M$, and the original input latent $z_f^T$. 
The input appearance and motion condition are aligned with the corresponding reference and 3D mesh image. 
The generation process can be abbreviated as:
\begin{equation}
    z_f^{0} = \hat{X}_\theta^f(ControlN(p^f), c_{s_r}^f, z_f^{T}, M, z_{M}),
\end{equation}
where $\hat{X}_\theta^f$ is the face SGDM, $z_f^{0}$ is output, and $z_f^{T}$ is input latent.
The inpainting-style enhancement prevents the requirements of paired data and trains on only ground-truth.

\begin{algorithm}[t]
\caption{
Overlapped Temporal Denoising
}
\label{algorithm:temporal}
\KwIn{
Video Diffusion Model $\hat{X}_\theta^V$,
Timestep $t$,
Latent sequence $Z^t = \{z^t_1, z^t_2, ..., z^t_N\}$,
Pose sequence $\mathbb{P}=\{p_1, p_2, ..., p_N\}$,
Window size $ws$,
Overlap size $os$, 
$Scheduler$.
}
\KwResult{
Denoised latent sequence $Z^{t-1} = \{z^{t-1}_1, z^{t-1}_2, ..., z^{t-1}_N\}$
}
$count \leftarrow \{0,0,...,0\}$ \;
$Noise \leftarrow \{0,0,...,0\}$ \;
\For {$i = 0, ..., N/(ws-os)$}
{
    $wb \leftarrow i*(ws-os)$ \;
    $we \leftarrow i*(ws-os)+ws$ \;
    $Z^t_{wd}\leftarrow  Z^t[wb: we]$ \;
    $P_{wd}\leftarrow  P[wb: we]$ \;
    $N^t_{wd} \leftarrow \hat{X}_\theta^V(Z^t_{wd}, t, P_{wd})$ \;
    $Noise[wb: we] \leftarrow  Noise[wb: we] + N^t_{win}$ \;
    $count[wb: we] \leftarrow  count[wb: we]+1$ \;
    
}
\CommentSty{\# Normalize overlapped noises} \\
$Noise \leftarrow Noise / count $ \;
$Z^{t-1} \leftarrow Scheduler(Noise, t, Z^t)$ \;

\end{algorithm}

\section{Experiments}
\subsection{Experimental Settings}
\paragraph{Dataset.}
For pretraining, we follow \cite{cogesture_ginosar2019learningstylegesture, yi2023talkshow} to utilize 27 hours of video with SMPL-X annotation on four identities.
We apply robust video matting~\cite{lin2022robustmatting} on the pretraining videos to remove the background.
For one-shot video fine-tuning, we collect a dataset with ten identities from diverse sources, each with a one to five-minute video.
To validate the proposed method in various scenes, the ten identities are collected in three ways: four videos of the identities used for pertaining; three web videos of celebrities (Luo Xiang, Guo Degang, and one video from the YouTube channel ScienceOfPeople~\footnote{https://www.youtube.com/@ScienceOfPeople}); three footages of invited individuals using a green screen.
Videos are split into video clips, where each is $300$ frames, $30$ fps, and $10$ seconds long, surpassing the processing capability of existing video diffusion models.
We additionally collected some video clips for each identity to serve as their test motion samples.

\paragraph{Comparison Methods.}
We compare our method with four state-of-the-art approaches: Pose2Img~\cite{cogesture_qian2021speechdrivetemplates}, TPS~\cite{motiontransfer_zhao2022tpsmotion}, DreamPose~\cite{karras2023dreampose} and DisCo~\cite{wang2023disco}.
Pose2Img is part of the co-speech work SpeechDrivesTemplates~\cite{cogesture_qian2021speechdrivetemplates}, is a modified version of \cite{motiontransfer_balakrishnan2018warpingbyICCV2021} and achieves person-specific human video generation conditioned on human landmarks, which is the most relevant GAN-based method to ours.
TPS~\cite{motiontransfer_zhao2022tpsmotion} is a generic motion-transfer method modeling motion via thin plate spline transformation.
DreamPose~\cite{karras2023dreampose} and DisCo~\cite{wang2023disco} are diffusion-based human video generation systems.
We train Pose2Img from scratch and fine-tune DreampPose and DisCo with our data while keeping the TPS's generic weight.

\paragraph{Implementation Details.}
We manually crop the body region into 512$\times$512 pixels, while face enhancement is 256$\times$256.
The pre-training stage with body generation lasted for 300K training steps with batch size 4, learning rate 1e-5, which took about seven days.
Fine-tuning on one identity takes about one day, with 50K training steps for body generation and 80K steps for face enhancement.
All experiments are operated on one Nvidia 40G A100 GPU.
For inference, CFG~\cite{ho2022cfg} is set as 7.5.
We found higher ControlNet condition scale $W_c$ benefits preserving the structure of 3D mesh and hand gestures, and we set $W_c$ as 2.
SMPL-X annotations are produced by the tools from \cite{yi2023talkshow}.
For face enhancement, FFHQ~\cite{karras2019stylegan} preprocess method is utilized to align and crop the face images, and the inpainting mask is extracted by a facial segment model.
For overlapped temporal denoising, the window size $ws$ is set to $16$, and the overlap size $os$ is four.
The number of denoising inference steps is set to 20.

\begin{figure*}
  \centering
   \includegraphics[width=1.0\linewidth]{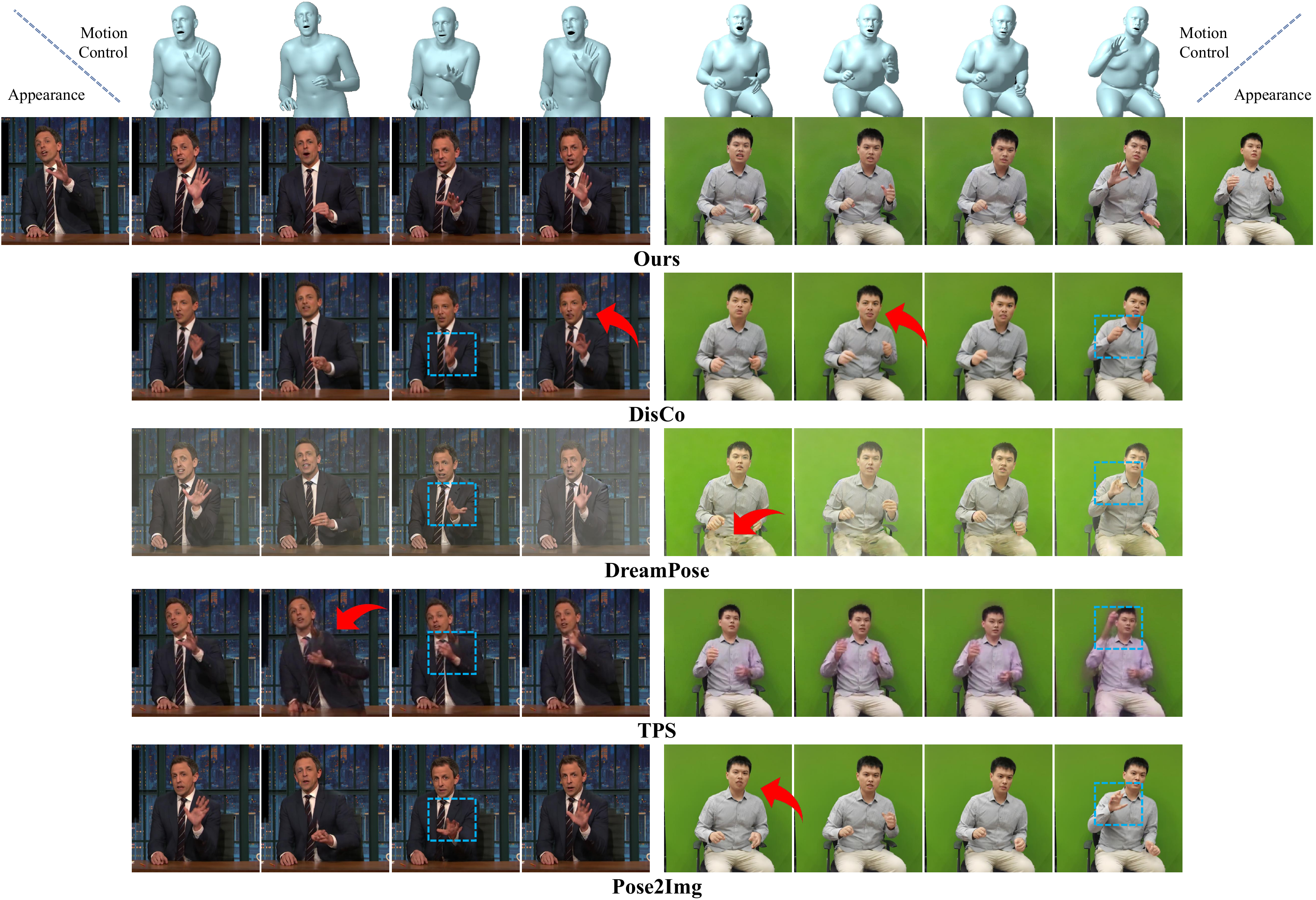}
   \caption{
   Qualitative results compared with other methods. Our methods achieve accurate gestures and high-quality generation with facial details. More results are provided in supplementary materials.
   }
  \label{fig:samidmotion}
\end{figure*}

\paragraph{Metrics.}
We use FID~\cite{heusel2017fid} to measure image quality, FVD~\cite{unterthiner2018fvd} to measure temporal consistency.
Landmark Mean Distances (LMD) are performed to calculate the preservation of body structure between the video frame and input 3D mesh.
We compute the LMD on the face, body, and hands using OpenPose~\cite{cao2017openpose}.
For the inaccuracy of landmarks prediction on hands, the outliers are ignored during computing LMD.

\subsection{Main Results}
\label{sec:expr.main_results}
We utilize the data mentioned above as our experiment dataset, where 10 identities from different sources are split into 9:1 as train and test data.
Our experiment is set with 30 video clips as a test, as each identity takes 3 test data.
Audio-driven results are further provided in supplementary materials, where we generate motions by TalkSHOW~\cite{yi2023talkshow}.

\paragraph{Quantitative Results.}
Quantitative results are presented in Table~\ref{tab:main}.
Our approach achieves the best results on FID and FVD, representing the image quality and video temporal consistency.
Ours get the highest results on face and hand, while comparable to other methods.
Pose2Img
It is worth noting that hand landmarks are not accurate for complex gestures, which is not sufficient to reflect the performance. 
Besides, the landmarks used to compute LMD are also used for Pose2Img and DisCo inference, which causes their results to be somewhat biased.
We conducted a user study in supplementary materials to provide complementary quantitative results.

\begin{table}
  \centering
  \resizebox{\linewidth}{!}{
  \begin{tabular}{@{}lcccccc@{}}
    \toprule
    Method                                    & FID↓              & FVD↓           &    LMD (Face)↓    & LMD (Body)↓   & LMD (Hand)↓               \\
    \midrule
    Pose2Img                                  &  51.92            & 328.18         &    3.73           & \textbf{4.78} &  6.96              \\
    TPS                                       &  136.02           & 884.97         &    5.22           & 8.37          &  18.72             \\
    DreamPose                                 &  83.47            & 868.20         &    4.21           & 5.92          &  13.85             \\
    DisCo                                     &  60.95            & 390.77         &    4.15           & 5.11          &  11.21                \\
    Ours                                      &  \textbf{40.33}   & \textbf{139.82}&    \textbf{1.44}  & 4.88          &  \textbf{5.41}     \\
    \midrule
     w/o TD                                   &       48.84       &  344.85        &   1.45            &  4.74         &  5.61              \\
    w/o FE                                    &       40.84       &  124.10        &   1.56            &  -            &   -                \\

    w/o Two-Stage Setting I                                 &      55.87        &  278.73        &    4.38          &  6.92    &   7.25                \\
    w/o Two-Stage Setting II                                &      53.99        &  178.79        &    1.56          &  4.96    &   6.01                \\
    SMPL perturbation                          &  39.56            &  136.21        &    1.40          &  4.33    &  5.25                 \\
    
    
    \bottomrule
  \end{tabular}
  }
  \caption{Quantitative results of our method compared with SOTAs and ablation studies. Our method achieves better performance on image quality, temporal consistency, and structure preservation.}
  \label{tab:main}
  \vspace{-5pt}
\end{table}

\paragraph{Qualitative Results.}
The qualitative results of our experiment are shown in Fig.~\ref{fig:samidmotion}.
In this figures, the blue boxes indicate distorted hand structures, and arrows label appearance artifacts.
Our methods accomplish the best image quality and appearance preservation, and simultaneously faithfully reflect mouth movements and hand gestures.
DisCo achieves high-quality generation through diffusion methods, yet it lacks focus on generating intricate details for hands and faces, resulting in noticeable imperfections within our specific context.
DreamPose focuses on generating videos within simple scenes, such as fashion-related contexts. 
In our scenario, it faces notable challenges in generating backgrounds and handling complex actions.
TPS is a general warping-based method where, in unfamiliar domains, its generated poses are challenging to surpass beyond the given reference image's pose.
As for a person-specific method, Pose2Img achieves better results compared to another comparison, while the artifacts are apparent in hands and lips because of the limitation of warping-based methods.
Our approach learns the mapping from motion to appearance via a diffusion model, complemented by a temporal scheme, enabling the generation of sufficiently high-quality human videos.

\begin{figure}[htp]
\vspace{-10px}
\newcommand{\galleryfigurewidth}{0.10}
    \begin{minipage}[t]{0.32\linewidth}
        \includegraphics[width=\linewidth]{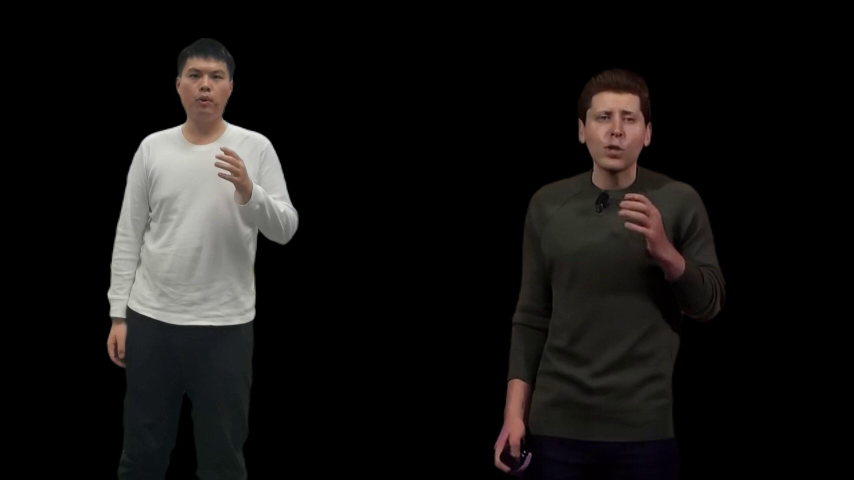}
    \end{minipage}
    \begin{minipage}[t]{0.32\linewidth}
        \includegraphics[width=\linewidth]{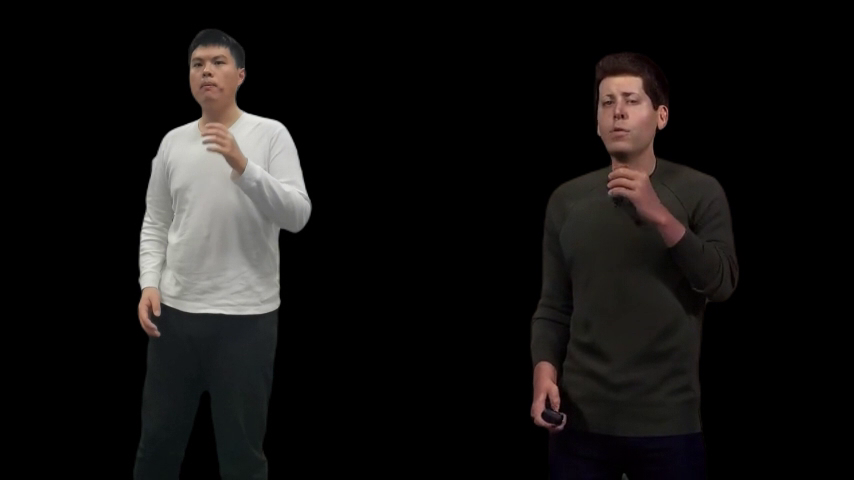}
    \end{minipage}
    \begin{minipage}[t]{0.32\linewidth}
        \animategraphics[width=\linewidth,autoplay=True]{30}{gifs/crossmotion/000}{183}{210}
    \end{minipage}
\caption{
    Cross-person motion results. 
    For each image, left is pose, and right is output. 
    Click the last images to play the embedded clips with Acrobat Reader.
}
\label{fig:crossperson}
\end{figure}

\begin{figure}[htp]
\vspace{-10px}
\newcommand{\galleryfigurewidth}{0.10}
    \begin{minipage}[t]{0.32\linewidth}
        \includegraphics[width=\linewidth]{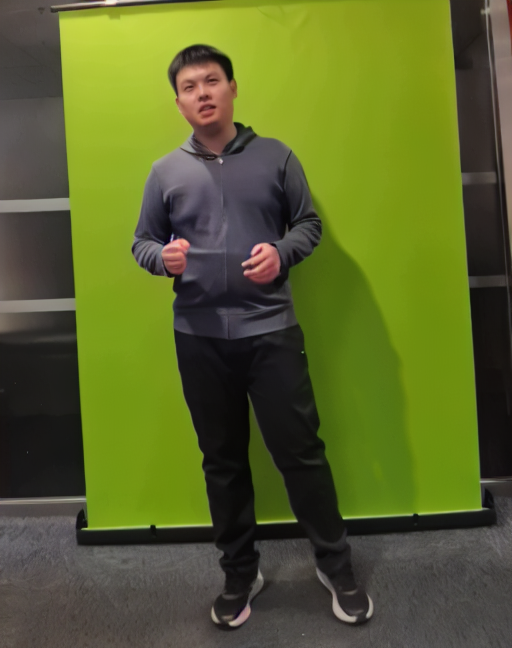}
    \end{minipage}
    \begin{minipage}[t]{0.32\linewidth}
        \includegraphics[width=\linewidth]{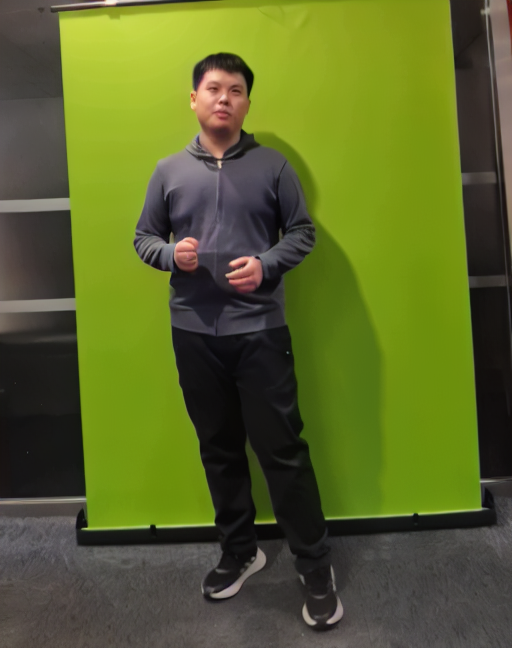}
    \end{minipage}
    \begin{minipage}[t]{0.32\linewidth}
        \animategraphics[width=\linewidth,autoplay=True]{30}{gifs/fullbody/000}{281}{292}
    \end{minipage}
\caption{
    Full-body results. Click the last images to play the embedded clips with Acrobat Reader.
}
\label{fig:fullbody}
\end{figure}

\paragraph{More results.} 
We further provide cross-person motion and full-body results.
The cross-person motion-driven results are shown in Fig.~\ref{fig:crossperson}.
When the driven motion is in a similar style as the target person (a standing man to a standing man), the generated results are also in good visual quality.
Full-body results are provided in Fig.~\ref{fig:fullbody}.
Although the full-body talking style differs from our training set, the results are still of good quality.

\subsection{Ablation Study}
\paragraph{Validation of Batch-overlapped Temporal Denoising.}
To verify the effectiveness of Batch-overlapped Temporal Denoising (TD), we conduct analyses from multiple aspects. The numerical results of ablation are displayed in Table~\ref{tab:main}, where without TD, our method's performance decreases significantly. We still get better results on temporal measures compared with DisCo because of the detailed pose condition and performs slightly worse compared to warping-based methods that leverage inter-frame advantages.
The LMDs are not influenced much, as the generated structure is more influenced by SGDM. 

Besides, we analyse the hyperparameters of TD, the window size $wd$ and overlapped size $os$. The default setting is $ws=16, os=4$. we vary them as $ws=8, os=2$ and $ws=32, os=8$. The numerical results are displayed in Table~\ref{tab:ablation.temporal}, where temporal-related measures, FVD improve with an increase in settings. We select $ws=16, os=4$ as the inference cost of the all-frame cross-attention module is increased as $O(ws^2)$.

\begin{figure}[htp]
\newcommand{\galleryfigurewidth}{0.24}
\centering
\begin{subfigure}[t]{\galleryfigurewidth\linewidth}
    \subcaption*{Frame1}
\end{subfigure}
\begin{subfigure}[t]{\galleryfigurewidth\linewidth}
    \subcaption*{Frame2}
\end{subfigure}
\begin{subfigure}[t]{\galleryfigurewidth\linewidth}
    \subcaption*{Frame3}
\end{subfigure}
\begin{subfigure}[t]{\galleryfigurewidth\linewidth}
    \subcaption*{Frame4}
\end{subfigure}
    \begin{minipage}[t]{\linewidth}
    \centering
        \begin{minipage}{\galleryfigurewidth\linewidth}
            \includegraphics[width=\linewidth]{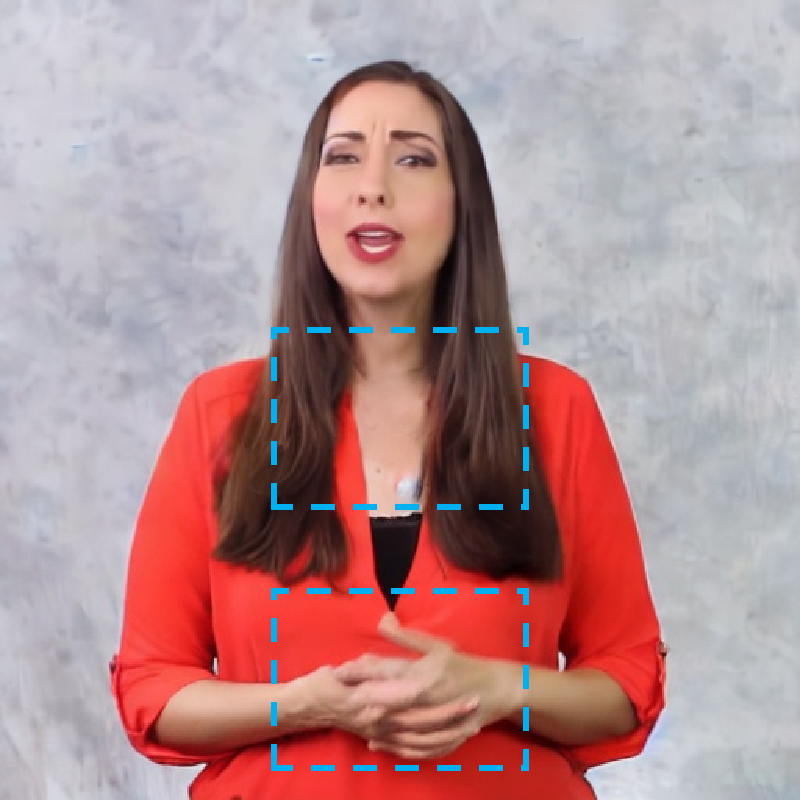}
        \end{minipage}
        \begin{minipage}{\galleryfigurewidth\linewidth}
            \includegraphics[width=\linewidth]{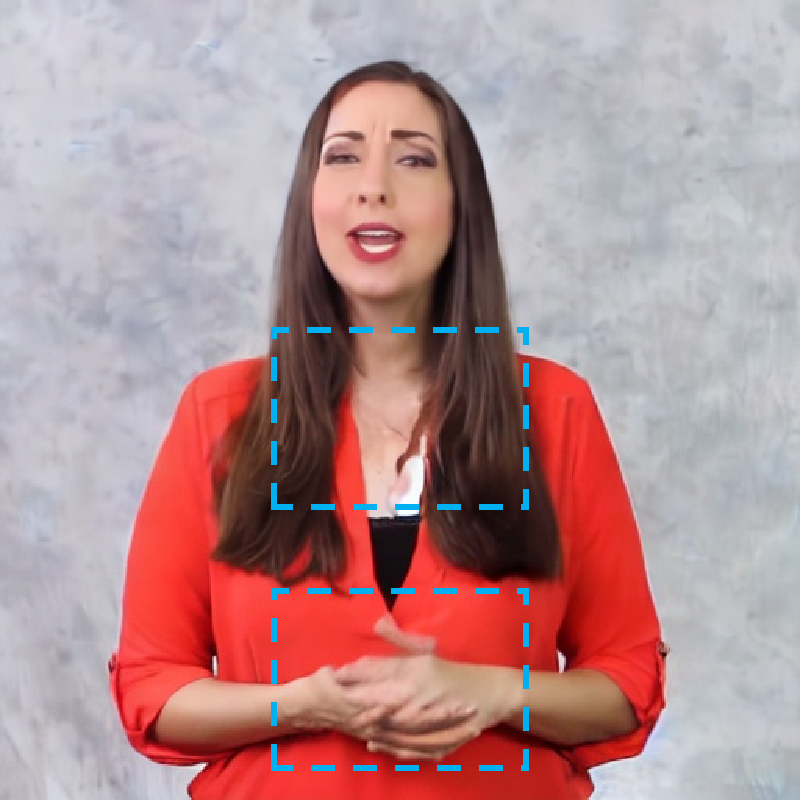}
        \end{minipage}
        \begin{minipage}{\galleryfigurewidth\linewidth}
            \includegraphics[width=\linewidth]{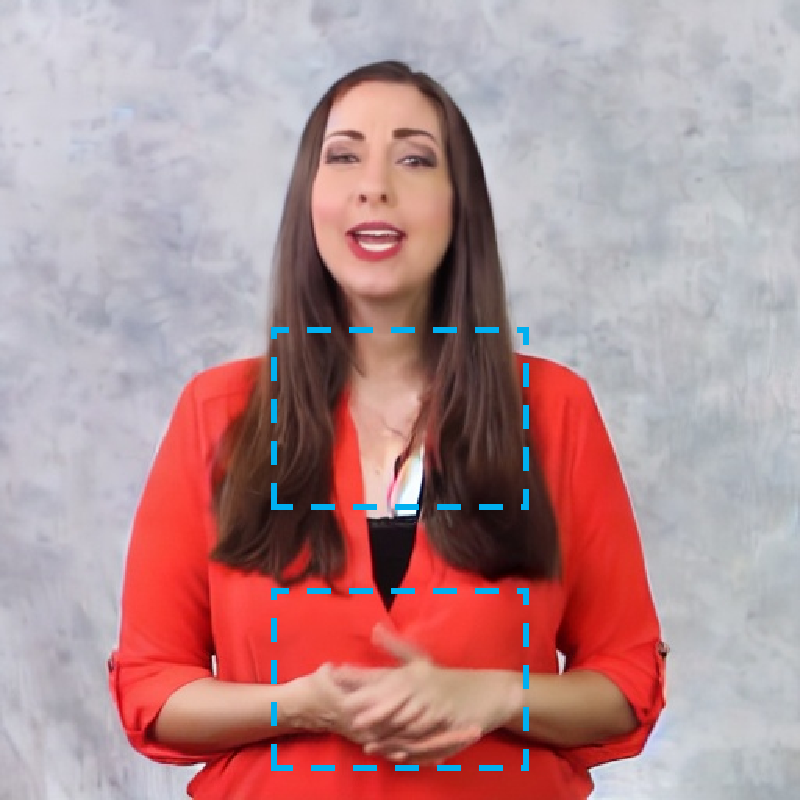}
        \end{minipage}
        \begin{minipage}{\galleryfigurewidth\linewidth}
            \includegraphics[width=\linewidth]{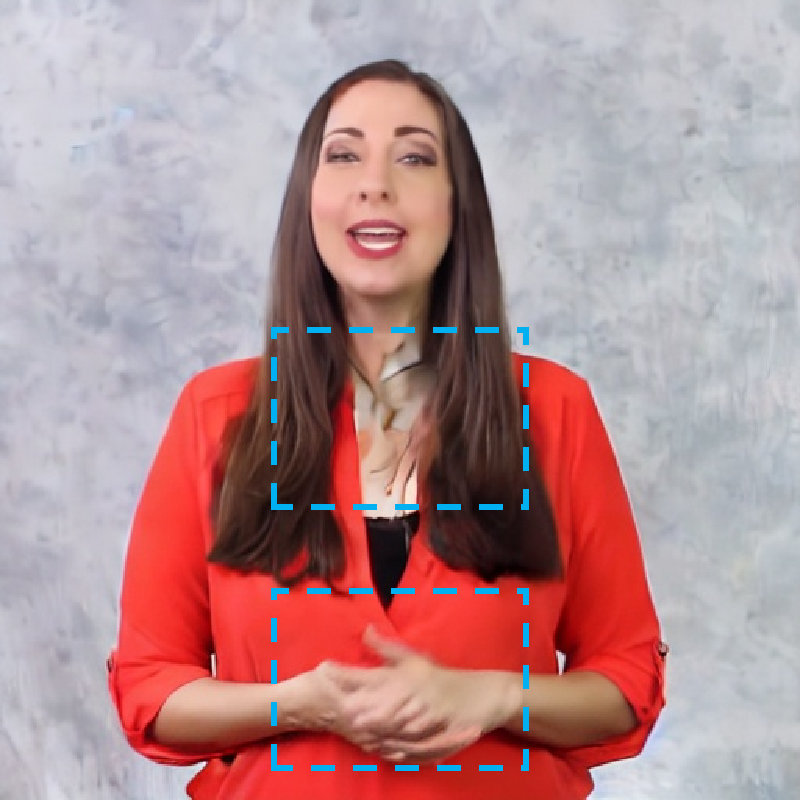}
        \end{minipage}
        \subcaption{W/o batch-overlapped temporal denoising}
    \end{minipage}

  \begin{minipage}[t]{\linewidth}
    \centering
        \begin{minipage}{\galleryfigurewidth\linewidth}
            \includegraphics[width=\linewidth]{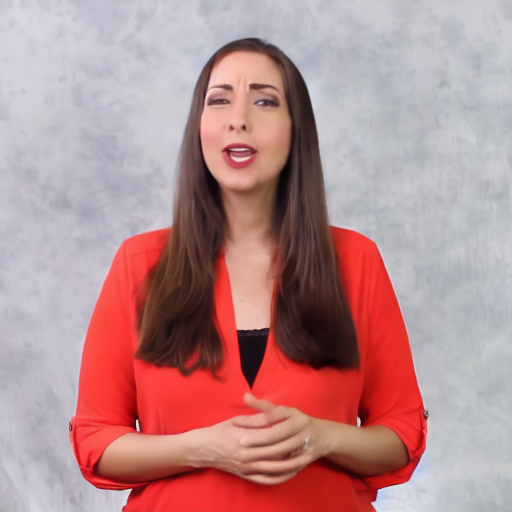}
        \end{minipage}
        \begin{minipage}{\galleryfigurewidth\linewidth}
            \includegraphics[width=\linewidth]{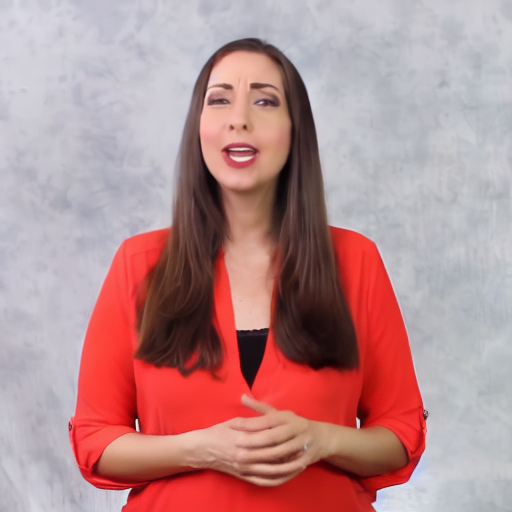}
        \end{minipage}
        \begin{minipage}{\galleryfigurewidth\linewidth}
            \includegraphics[width=\linewidth]{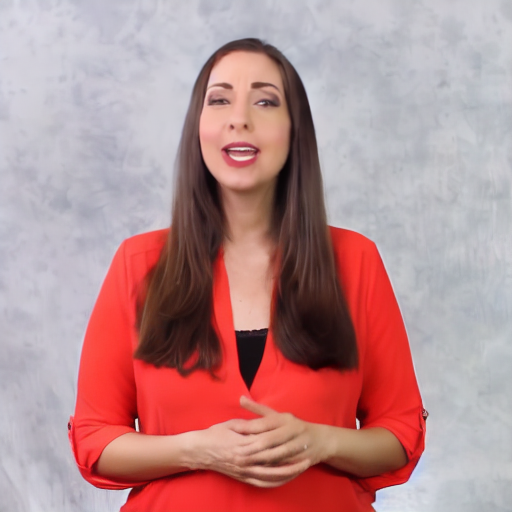}
        \end{minipage}
        \begin{minipage}{\galleryfigurewidth\linewidth}
            \includegraphics[width=\linewidth]{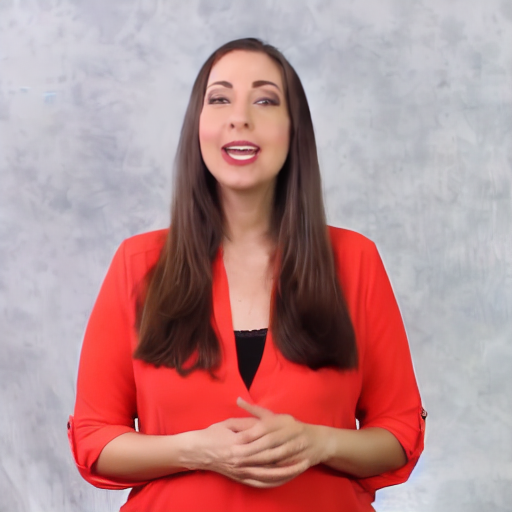}
        \end{minipage}
        \subcaption{With batch-overlapped temporal denoising }
    \end{minipage}


\caption{
  Ablation analysis of temporal denoising.
  The \textbf{first row} is the result without batch-overlapped temporal denoising, whereas the \textbf{second row} employed it.
  Without temporal denoising, the generated videos may exhibit discontinuities such as flickering and ghosting as the artifacts around the neck region on the first row.
}
\label{fig:ablation.temporal}
\end{figure}

Another analysis is the qualitative results, as shown in Fig.~\ref{fig:ablation.temporal}. We observed that without TD, the generated videos may exhibit temporal artifacts such as flickering and ghosting. 
This could be because the image model during training hasn't accurately captured the distribution of human bodies within the dataset, leading to some unforeseen patterns. Introducing contextual information by Batch-Overlapped Temporal Denoising could better eliminate these unintended artifacts.

\begin{figure}[htp]
\newcommand{\galleryfigurewidth}{0.30}
\centering
    \begin{minipage}[t]{\linewidth}
    \centering
        \begin{minipage}{\galleryfigurewidth\linewidth}
            \includegraphics[width=\linewidth]{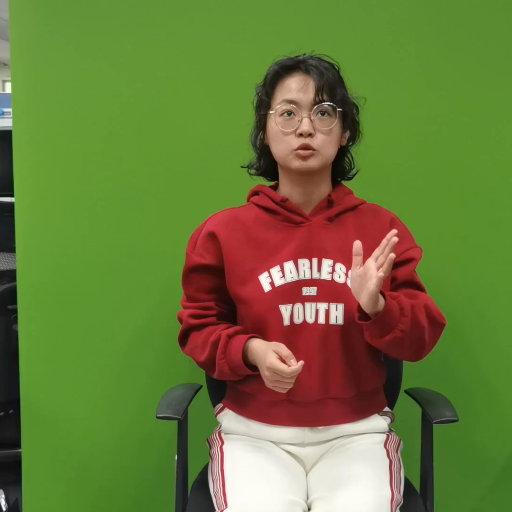}
            \includegraphics[width=\linewidth]{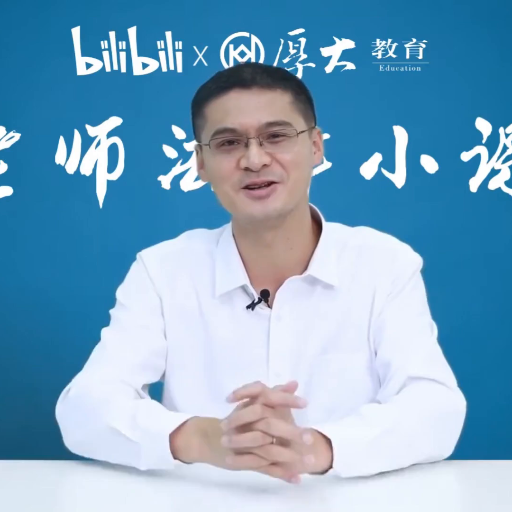}
        \end{minipage}
        \begin{minipage}{\galleryfigurewidth\linewidth}
            \includegraphics[width=\linewidth]{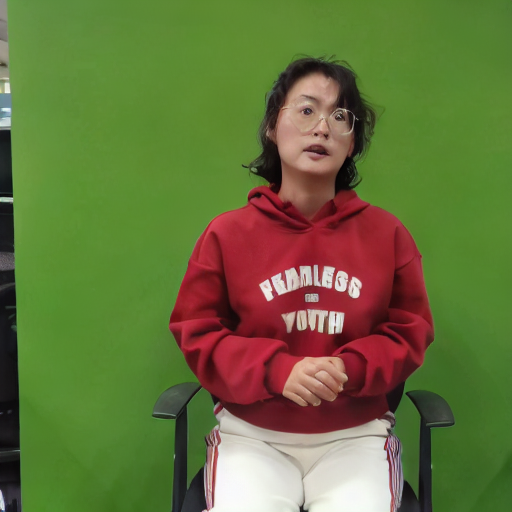}
            \includegraphics[width=\linewidth]{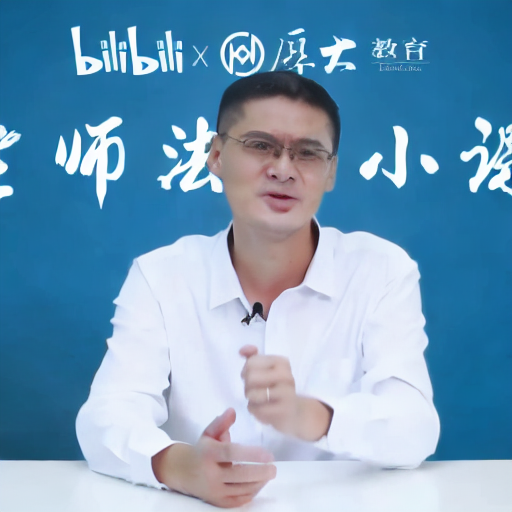}
        \end{minipage}
        \begin{minipage}{\galleryfigurewidth\linewidth}
            \includegraphics[width=\linewidth]{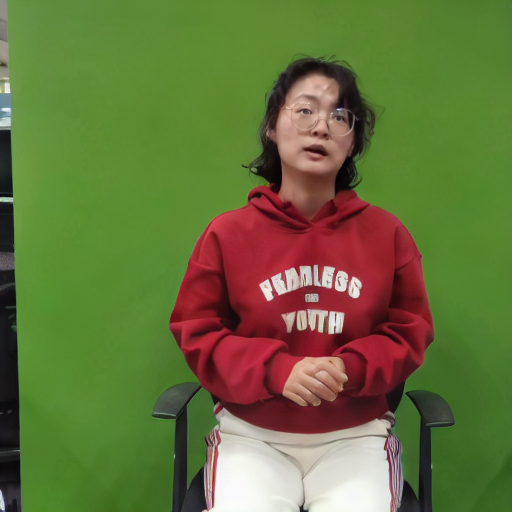}
            \includegraphics[width=\linewidth]{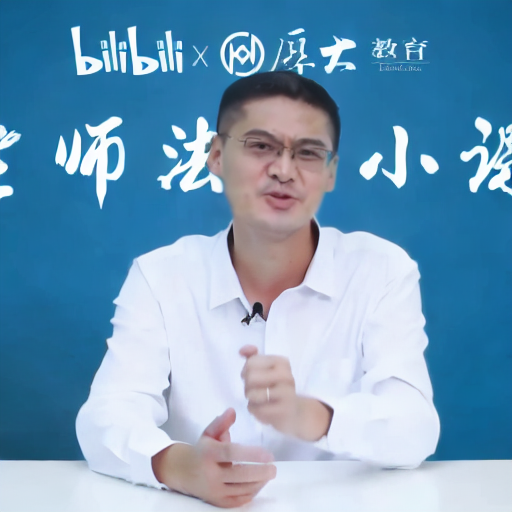}
        \end{minipage}

    \end{minipage}
\begin{subfigure}[t]{\galleryfigurewidth\linewidth}
    \subcaption*{Appearance}
\end{subfigure}
\begin{subfigure}[t]{\galleryfigurewidth\linewidth}
    \subcaption*{W/o face enh.}
\end{subfigure}
\begin{subfigure}[t]{\galleryfigurewidth\linewidth}
    \subcaption*{With face enh.}
\end{subfigure}
\caption{
Ablations of identity-specific face enhancement.
Directly training models to generate entire human body images often leads to the creation of facial features with poor realism. Our identity-specific face enhancement significantly improves the quality of generated content in the face region.
}
\label{fig:ablation.face}
\end{figure}

\paragraph{Validation of Identity-Specific Face Enhancement.}
We take an analysis of the Identity-Specific Face Enhancement (FE) module. The quantitative results of the ablation study on this module are shown in Table~\ref{tab:main}. 
LMD (Face) is improved by our face enhancement results, which reveals the capability of FE to generate facial detail. Besides, FVD result is slightly lower down due to that we utilize FE without temporal denoising.
The qualitative results of FE are shown in Fig.~\ref{fig:ablation.face}. We can distinctly observe that similar to other diffusion-based methods explicitly shown in Fig.~\ref{fig:samidmotion}, our approach sometimes struggles to directly generate high-quality facial features. FE can significantly improve facial details.

\begin{table}
  \centering
  \resizebox{\linewidth}{!}{
  \begin{tabular}{@{}lcccccc@{}}
    \toprule
    Method          &   FID↓            & FVD↓             & LMD (Face)↓       & LMD (Body)↓    & LMD (Hand)↓     \\
    \midrule
    $ws=8, os=2$      &  39.64          &        140.29    &    1.45           & 4.86          &  5.46                                        \\
    $ws=16, os=4$     &  40.33          &         139.82   &    1.44           & 4.88          &  5.41              \\
    $ws=32, os=8$     &  41.01          &       138.48     &    1.44           & 4.89          &  5.33                             \\
    \bottomrule
  \end{tabular}}
  \caption{ Analysis of varying the hyperparameters of batch-overlapped temporal denoising, window size $ws$ and overlapped size $os$. The default setting of our method is $ws=16, os=4$.}
  \label{tab:ablation.temporal}
\end{table}

\begin{figure}[t]
\vspace{-10px}
\newcommand{\galleryfigurewidth}{0.10}

    \begin{minipage}[t]{0.14\linewidth}
        \includegraphics[width=\linewidth]{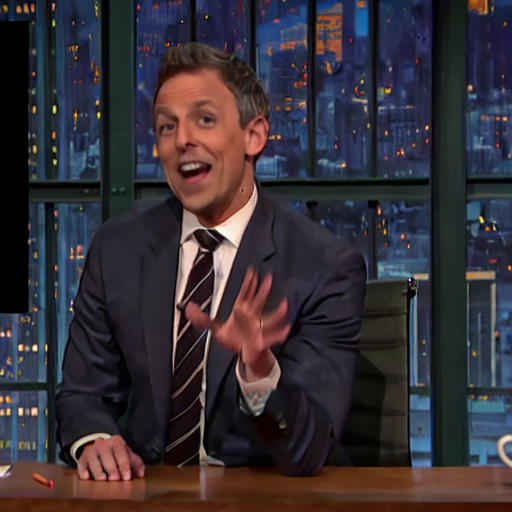}
    \end{minipage}
    \begin{minipage}[t]{0.14\linewidth}
        \includegraphics[width=\linewidth]{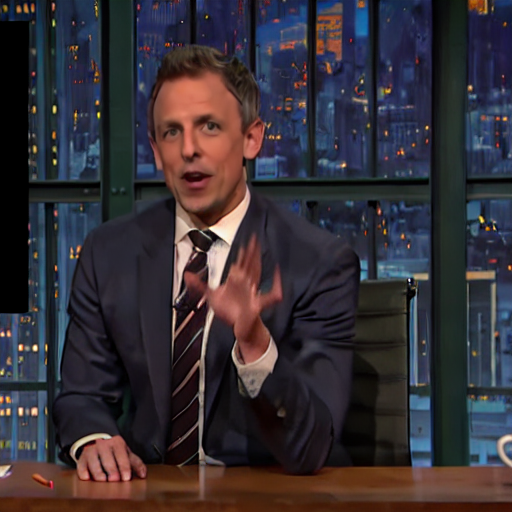}
    \end{minipage}
    \begin{minipage}[t]{0.14\linewidth}
        \animategraphics[width=\linewidth,autoplay=True]{30}{gifs/twostage/nopretrain/0000}{76}{88}
    \end{minipage}
    \begin{minipage}[t]{0.14\linewidth}
        \includegraphics[width=\linewidth]{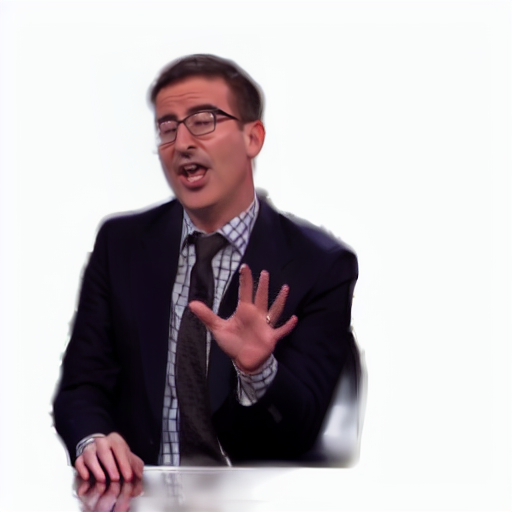}
    \end{minipage}
    \begin{minipage}[t]{0.14\linewidth}
        \includegraphics[width=\linewidth]{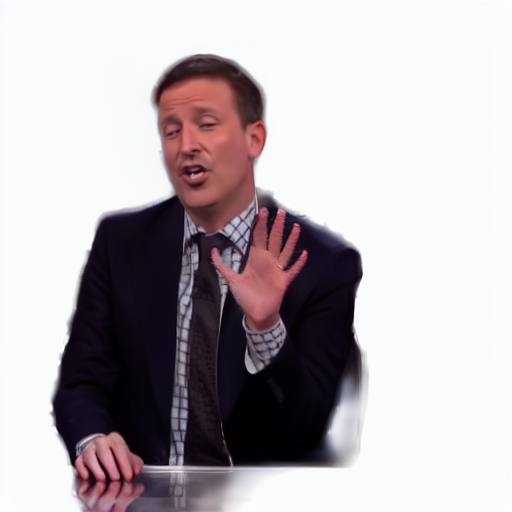}
    \end{minipage}
    \begin{minipage}[t]{0.14\linewidth}
        \animategraphics[width=\linewidth,autoplay=True]{30}{gifs/twostage/alltrainonce/0000}{76}{88}
    \end{minipage}
    \begin{subfigure}[t]{0.48\linewidth}
        \vspace{-15px}
        \subcaption*{\small (a) Setting I.}
    \end{subfigure}
    \begin{subfigure}[t]{0.48\linewidth}
        \vspace{-15px}
        \subcaption*{\small (b) Setting II.}
    \end{subfigure}
\vspace{-10px}
\caption{
Validation of two-stage training. 
Click the last images to play the embedded clips with Acrobat Reader.
}
\label{fig:ablation.twostage}
\end{figure}

\paragraph{Validation of Two-Stage Training.}
Without two-stage training, there are two alternative approaches. 
Setting I: directly training our model with videos of the target subject; Setting II: training target with all other subject data.
As shown in Fig.~\ref{fig:ablation.twostage}~(a) and Table~\ref{tab:main}, the result of Setting I lacks enough training data for pose-control ability.
As shown in Fig.~\ref{fig:ablation.twostage}~(b) and Table~\ref{tab:main}, the appearance and identity may mix with other subjects for Setting II. 
Meanwhile, it is hard to extend to a new subject for such a strategy.
With two-stage training, our model can control the pose and preserve the appearance of the target subject.

\begin{figure}[htp]
\vspace{-10px}
\newcommand{\galleryfigurewidth}{0.10}
    \begin{minipage}[t]{0.32\linewidth}
        \includegraphics[width=\linewidth]{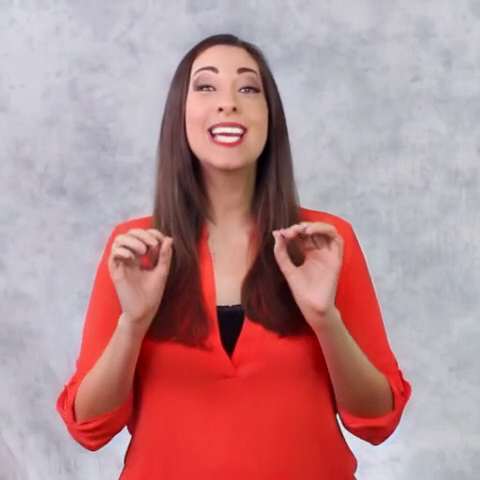}
    \end{minipage}
    \begin{minipage}[t]{0.32\linewidth}
        \includegraphics[width=\linewidth]{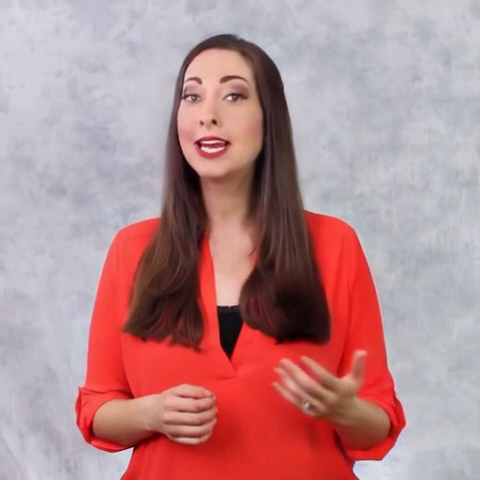}
    \end{minipage}
    \begin{minipage}[t]{0.32\linewidth}
        \animategraphics[width=\linewidth,autoplay=True]{30}{gifs/perturb/000}{074}{99}
    \end{minipage}
\caption{
    Trained with SMPL perturbation. 
    Click the last images to play the embedded clips with Acrobat Reader.
}
\label{fig:ablation.smplxparam}
\end{figure}

\paragraph{Validation of SMPL-X Parameters.}
Estimating SMPL inevitably comes with errors. We preprocess the extracted SMPL following TalkShow~\cite{yi2023talkshow}, optimizing SMPL through multiple estimations to reduce errors.
We further explore imperfect SMPL by perturbing the SMPL parameters during training.
Table~\ref{tab:main} and Fig.~\ref{fig:ablation.smplxparam} show that our method can still obey the input pose. The better statistics indicate that SMPL perturbing, serving as a kind of data augmentation, could further improve the performance. 

\section{Conclusion}
In this paper, we propose ``Make-Your-Anchor'', a diffusion-based 2D avatar generation framework to produce realistic and high-quality anchor-style human videos.
In this framework, frame-wise motion-to-appearance diffusing is proposed to train a structure-guided diffusion model with a two-stage training strategy and accomplish binding specific appearance with movements via a binding style approach.
To generate temporal consistent human video, we propose a training-free strategy to extend the image diffusion model into a video diffusion model with an all-frame cross-attention module and design a batch-overlapped temporal denoising algorithm to overcome the limitation on the generated video length.
Identity-specific face enhancement is introduced 
from the observation that face details are difficult to reconstruct during holistic human generation. An inpainting-style enhancement module is applied to the human image to solve this problem.
Through the amalgamation of our entire system approach, our framework successfully generates high-quality, structure-preserving, and temporal coherence anchor-style human videos, which may offer some reference value for the widely applicable technology of 2D digital avatars.

\noindent \paragraph{Limitation and further work.}
\label{sec:limitation}
Despite the capability of our method to produce high-quality videos, if the human body orientation for inference significantly differs from what has been observed in the fine-tuning videos, there may be issues with preserving the appearance. This occurs due to the binding style training. 
From the other perspective, an increase in the quantity of pre-training and fine-tuning data may handle the generation of more complex movements and orientations.
Another limitation is that our method does not model foregrounds with occlusions, potentially resulting in the ghosting of occluded elements. One possible solution could be segmenting occluded elements as a reference to enable the model to preserve the occlusions.

\section*{Acknowledgements}
This work was partly supported by the National Natural Science Foundation of China under No. 62102162, Beijing Science and Technology Plan Project under No. Z231100005923033, and the National Science and Technology Council under No. 111-2221-E-006-112-MY3, Taiwan.

{
    \small
    \bibliographystyle{ieeenat_fullname}
    \bibliography{main}
}


\end{document}